\documentclass{article} 
\usepackage{mathptmx} % This is Times font
\newcommand{\ignore}[1]{}
\usepackage{fancyhdr}
\usepackage[normalem]{ulem}
\usepackage[hyphens]{url}
\usepackage{microtype}
\usepackage{array}
\usepackage{multirow}% http://ctan.org/pkg/multirow
\usepackage{hhline}
\usepackage{graphicx}
\usepackage{subcaption}
\usepackage{booktabs}
\usepackage{amsmath}
\usepackage{soul}
\usepackage{multirow}

\usepackage[accepted]{icml2019}

\usepackage[bookmarks=true,breaklinks=true,colorlinks,linkcolor=black,citecolor=blue,urlcolor=black]{hyperref}

\icmltitlerunning{Optimizing Multi-GPU Parallelization Strategies}

\begin{document}
\sloppy
\twocolumn[
\icmltitle{{Optimizing Multi-GPU Parallelization Strategies for Deep Learning Training}}
%\vskip 0.3in
%\vspace{-0.5cm}

%\icmlsetsymbol{go}{*}

\begin{icmlauthorlist}
\icmlauthor{Saptadeep Pal}{go}
\icmlauthor{Eiman Ebrahimi}{do}
\icmlauthor{Arslan Zulfiqar}{do}
\icmlauthor{Yaosheng Fu}{do}\\
\icmlauthor{Victor Zhang}{do}
\icmlauthor{Szymon Migacz}{do}
\icmlauthor{David Nellans}{do}
\icmlauthor{Puneet Gupta}{go}
\end{icmlauthorlist}
\vspace{0.3cm}

\icmlaffiliation{go}{University of California Los Angeles}
\icmlaffiliation{do}{NVIDIA\\}
\icmlcorrespondingauthor{Eiman Ebrahimi}{eebrahimi@nvidia.com}
\icmlcorrespondingauthor{Saptadeep Pal}{saptadeep@ucla.edu}

]
\printAffiliationsAndNotice{} % otherwise use the standard text.

\begin{abstract}
Deploying deep learning (DL) models across multiple compute devices to train large and complex models continues to grow in importance because of the demand for faster and more frequent training. Data parallelism (DP) is the most widely used parallelization strategy, but as the number of devices in data parallel training grows, so does the communication overhead between devices. Additionally, a larger aggregate batch size per step leads to statistical efficiency loss, i.e., a larger number of epochs are required to converge to a desired accuracy. These factors affect overall training time and beyond a certain number of devices, the speedup from leveraging DP begins to scale poorly. In addition to DP, each training step can be accelerated by exploiting model parallelism (MP). 
%However, the inherent parallelism of a model, or its implementation, can limit the number of devices the model scales to using MP alone. 
%EIMAN-1-23-19: The above sentence adds no value to the abstract. It probably never did since we moved introducing model parallelism out of the abstract which was the correct thing to do. Reads much better without it, please do not re-introduce.
This work explores hybrid parallelization, where each data parallel worker is comprised of more than one device, across which the model dataflow graph (DFG) is split using MP. We show that at-scale, hybrid training will be more effective at minimizing end-to-end training time than exploiting DP alone.  We project that for Inception-V3, GNMT, and BigLSTM, the hybrid strategy provides an end-to-end training speedup of at least 26.5\%, 8\%, and 22\% respectively compared to what DP alone can achieve at scale.
%We evaluate hybrid training's effectiveness on three DL networks and conservatively project that on average, it can perform at least 15\% better than DP alone.

\end{abstract}
\vspace{-0.7cm}
\section{Introduction}

Deep learning (DL) models continue to grow and the datasets used to train them are increasing in size, leading to longer training times. Therefore, training is being accelerated by deploying DL models across multiple devices (e.g., GPUs/TPUs) in parallel. Data parallelism (DP) is the simplest parallelization strategy~\cite{Krizhevsky_2017,Jeff_2012,Karen_2014}, where replicas of a model are trained on independent devices using independent subsets of data, referred to as mini-batches. All major frameworks (e.g. TensorFlow~\cite{tensorflow}, PyTorch~\cite{pytorch}) support DP using easy-to-use and intuitive APIs~\cite{horovod}. However, as the number of devices used to exploit DP increases, the global batch size also typically increases~\footnote{We discuss our methodology in Section~\ref{sec:measuring_epoch} and other possibilities in Section~\ref{sec:related_work}.}. This poses a fundamental problem for data parallel scalability because for any given DL network, there exists a global batch size beyond which converging to the desired accuracy requires a significantly larger number of iterations.
%a significantly increased number of iterations are required to converge to the desired accuracy. 
This is primarily due to the reduced statistical efficiency of the training process~\cite{hoffer2017train}. In addition, as the number of devices employed increases, the synchronization/communication overhead of sharing gradients across devices increases, further limiting overall training speedup. 

Model parallelism (MP) is a complementary technique in which the model dataflow graph (DFG) is split across multiple devices while working on the same mini-batch~\cite{Jeff_2012, Seide_2014}.
MP has been traditionally used to split large models (which can not fit in a single device's memory), but employing MP can also help speed up each training step by placing and running concurrent operations on separate devices. Unfortunately the amount of parallelism that exists in today's models is often limited~\cite{wu_2016, szegedy_2015}, either by the algorithm or by its implementation. Therefore, using MP {\em alone} to obtain performance through parallelization typically does not scale well to a large number of devices. Additionally, maximizing the speedup from MP is often non-trivial~\cite{azalia_2017,mirhoseini2018a}. Optimizing MP requires carefully splitting the model to take into account the overhead of communicating activations (during the forward pass) and gradients (during the backward pass) between dependent operations (placed on separate devices) in order to achieve the maximum possible speedup.

This work studies which parallelization strategies to adopt to minimize end-to-end training time for a given DL model on available hardware. We ask the question: how can we improve the scaling obtained from DP, by combining MP and DP to achieve the best possible end-to-end training time at a given accuracy? The novel insight of this work is that when the number of devices (and hence global batch size) grows to a point where scaling from DP slows significantly, MP should then {\em be used in conjunction} with DP to continue improving training times. The speedup obtainable via MP is critical to this tipping point. We show that every network will have a unique scale at which DP's scaling and statistical efficiency degradation can be overcome by MP's speedup. 
%Because the training speedup obtainable via MP is critical to determining this tipping point, we show that every network will have a unique scale at which MP's speedup is large enough that it can overcome DP's scaling and statistical efficiency degradation. 
%EIMAN-1-23-19: the above sentence isn't great because it's not clear whether when we say "a unique scale" we mean that scale is referring to MP's speedup or DP's scaling and statistical efficiency degradation. In fact we're talking about the latter, but the former seems more likely from the structure of the sentence. 
This work makes the following contributions:
\vspace{-0.1cm}
\begin{itemize}
\setlength\itemsep{0em}
    \item
    We show that when DP's inefficiencies become large, a hybrid parallelization strategy where each parallel worker is model parallelized across multiple devices will further scale multi-device training.
    %The key factors inherently limiting end-to-end training speedup for DP are scaling efficiency and statistical efficiency loss. We show that when these inefficiencies become large, a hybrid parallelization strategy where each data parallel worker is model parallelized across multiple devices will further scale multi-device training.
    \item We develop an analytical framework to systematically find this cross-over point (in terms of number of devices - e.g., GPUs or TPUs - used to train a model) that indicates which parallelization strategy to use when optimizing training of a model, on a particular system.
    \item  We show that hybrid parallelization outperforms DP alone at different scales for different DL networks. We implement 2-way model parallel versions of Inception-V3, GNMT, and BigLSTM, and project that using them, hybrid training provides a speedup of at least 26.5\%, 8\%, and 22\% respectively above DP-only training at scale.
    %Finally we show that on three common DL networks, a reasonable crossover point (or number of devices) exists at which hybrid parallelization performs better than DP-alone. We project that on average, a hybrid strategy will perform at least 15\% better than DP-alone for large-scale training across these networks. 
    % EIMAN-1-23-19: I don't like *reasonable* at all in the above. 
    \item We propose $DLPlacer$, an integer linear programming based tool to find optimal operation-to-device placement to maximize MP speedup. We demonstrate DLPlacer's effectiveness by using it to derive an optimal placement for the Inception-V3 model~\cite{szegedy_2015}, showing the obtained 1.32x model parallel speedup with two GPUs is within 6\% of that predicted by the tool.
    
\end{itemize}

\vspace{-0.4cm}
\section{Background}
\label{sec:background}

\begin{figure}
    \centering
    \includegraphics[scale=0.45]{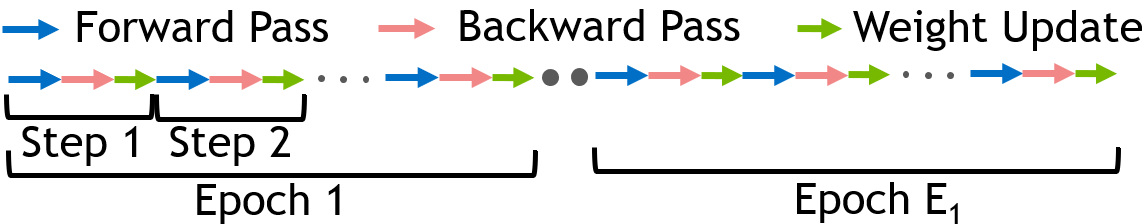}
    \caption{Deep learning training flow}
    \label{fig:training}
\end{figure}

{\em Neural Network Training:}
In neural network training, first a batch of inputs is forward propagated through the network to calculate the losses from each input. The losses are back propagated through the network to compute the gradients. The average of the batch's gradients is then used to update the weights. The size of the batch is chosen such that the compute resource of the device used for training is fully utilized. This process is called stochastic batch gradient descent~\cite{gardner_1984,martin_2010}. One forward and backward pass with gradient update to the weights is typically referred to as a training {\em step}. One iteration through the training data set where all inputs are processed once involves multiple steps and is referred to as an {\em epoch}. The training process, shown in Figure~\ref{fig:training} is run for multiple epochs until a desired training accuracy is reached.

\begin{figure}
    \centering
    \begin{subfigure}{0.5\textwidth}
        \centering
        \includegraphics[width=\linewidth]{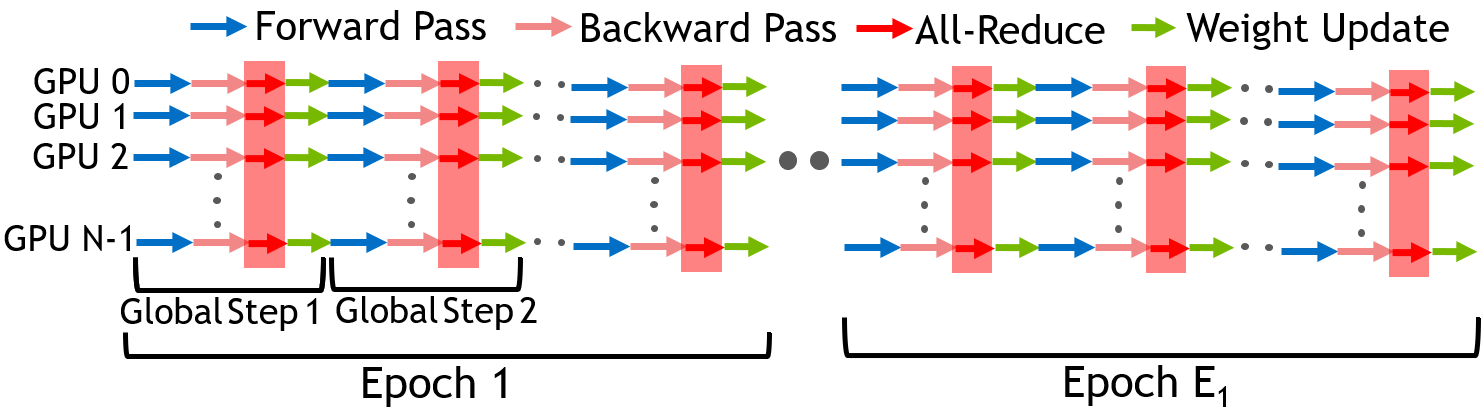}
        \caption{Data parallel training}
        \label{fig:data_parallel_training}
    \end{subfigure}
    
    \begin{subfigure}{0.5\textwidth}
        \centering
        \includegraphics[width=0.65\linewidth]{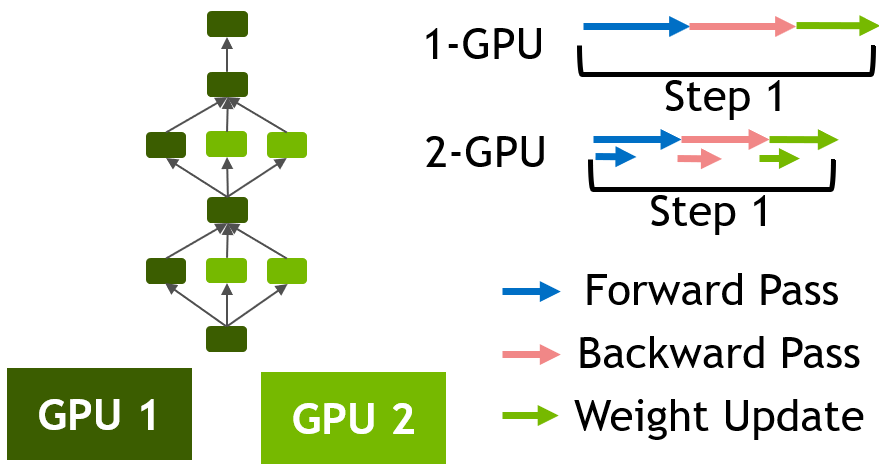}
        \caption{Model parallel training}
        \label{fig:model_parallel_training}
        \end{subfigure}
    \caption{Different Training Parallelization Strategies}
\end{figure}

{\em Data Parallel Training:} To accelerate training using DP,
a full set of model parameters (i.e., weights) are replicated across multiple devices/workers.
As Figure~\ref{fig:data_parallel_training} shows, each worker performs a forward and backward pass independently on a different batch of inputs first. Gradients are then communicated across workers and averaged; after which, each worker applies the same set of gradient values to the model weights. The communication of the gradients across the workers is done using {\em all-reduce} communication. 
The method of updating the model after each iteration (using the average of all gradients) is called synchronous stochastic gradient descent {\em(sync-SGD)} and is the most widely used technique for data parallel training. In this work, we call the batch of inputs per worker a {\em mini-batch} and the collection of all the mini-batches in a training step a {\em global batch}. 
%Thus, for a fixed mini-batch size, the global batch size increases proportionally with the number of data parallel workers.
%EIMAN-1-23-19: I don't think this last sentence is necessary here. If someone thinks that it is, lets talk about it. 

{\em Model Parallel Training:} The model DL is split by placing different operations of it's DFG onto different devices. This approach has been traditionally used for models whose parameters will not fit into a single device's memory~\cite{wu_2016, Krizhevsky_2017}. However, MP can provide per step training speedup~\cite{mirhoseini2018a, Jeff_2012} even when the entire model fits on one device by 
executing independent operations concurrently on separate devices, as shown in Figure~\ref{fig:model_parallel_training}. Splitting a DFG among multiple devices is non-trivial for many networks. The communication overhead of moving data between devices may be so large that it may outweigh the gains of MP. Thus, when dividing a network's DFG, characteristics such as compute intensity of each device, inter-device network bandwidth, and even network topology must be considered.

%{\em Operation Pipelining:} 
An alternative approach to obtaining speedup is to split up a model across multiple devices using pipelining~\cite{gpipe}. This enables splitting a model across multiple devices when a model does not have parallel branches and is sequential in nature. Networks are partitioned into groups containing one or a few layers of the network, where each group is placed on a different device. To orchestrate parallel execution, a mini-batch is split into yet smaller micro-batches and each device processes a different micro-batch sequentially but concurrently. While subtly distinct, for the purposes of this work we consider pipeline parallelism as an implementation instance of MP.

\vspace{-0.2cm}
\section{Decomposing End-to-End Training Time}
\label{sec:end_to_end}
End-to-end training time for a DL model depends on three factors: the average time per step ($T$), the number of steps per epoch ($S$) and the number of epochs ($E$) required to converge to a desired accuracy. Therefore, the total training time, i.e., time to converge ($C$) can be expressed as:

\vspace{-0.6cm}
\begin{equation}
\label{eq:training_speedup}
\begin{split}
    C = T \times S \times E
\end{split}
\end{equation}
\vspace{-0.6cm}

$T$ is determined by primarily by compute efficiency, i.e., given the same training setup, algorithm, and mini-batch size, $T$ depends solely on the compute capability of a device; better performing hardware provides smaller $T$ values. $S$ on the other hand, depends on the global batch size and number of items in the training dataset. All items in the dataset are processed once per epoch, therefore the number of steps per epoch ($S$) is equal to the number of items in the data set, divided by the {\em global batch size}. The number of epochs to converge ($E$) depends on the global batch size and other training hyper-parameters.

\subsection{Quantifying Data Parallel Training Time}\label{sec:speedup_dp}

In data parallel training, the network parameters (weights) are replicated across multiple worker devices and each worker performs a forward and a backward pass individually on a distinct batch of inputs (shown in Figure~\ref{fig:data_parallel_training}). \textcolor{black}{In this work, we focus on synchronous stochastic gradient decent ($sync-SGD$) for weight updates. In $sync-SGD$ workers are synchronized, i.e., the gradients from workers are shared and network parameters are updated such that all workers have the same parameters after each step. An alternative approach uses asynchronous updates, usually with a parameter server. When scaling to a large number of devices, this approach performs poorly~\cite{chen_2016}. Therefore, we use a ring-based all-reduce mechanism for data parallel training which provides superior performance and scalability over parameter server based approaches and primarily supports $sync-SGD$.}
We call the batch of inputs per worker a {\em mini-batch} and the collection of all the mini-batches in a training step a {\em global batch}.
When using DP alone to accelerate training, the speedup from employing N-way data parallelism ($SU_N$) compared to training {\em on a single device} can be expressed as:

\vspace{-0.5cm}
\begin{equation}
\label{eq:speedup_DP_1}
\begin{split}
    SU_N &= \frac{T_1}{T_N} \times \frac{S_1}{S_N} \times \frac{E_1}{E_N}
\end{split}
\end{equation}
\vspace{-0.5cm}

$T_1$ is the average training time per step when only one device is used for training, while $T_N$ is the time per step when $N$ data parallel devices (with a constant mini-batch size per device) are used. $T_N$ is always larger than $T_1$ because in DP, after each device has performed a forward and backward pass, the gradients must be exchanged between the devices using all-reduce communication (see Figure~\ref{fig:data_parallel_training})\footnote{Additionally, text and speech networks often exhibit straggler effects where processing some mini-batches take longer than others and therefore, in sync-SGD, devices with a shorter execution time of a mini-batch will suffer from under utilization~\cite{Hashemi_2018}}. Due to this communication overhead, $\frac{T_1}{T_N}$ will never be larger than one and is typically less than one. We call this ratio of $\frac{T_1}{T_N}$ the scaling efficiency ($SE_N$) of $N$-way DP.  

$S_1$ is the total number of steps required per epoch when one device is used, while $S_N$ is the number of steps per epoch when $N$ devices are used. When a single device is used, the global batch size is equal to the mini-batch size. In $N$-way data parallelism each device performs an independent step with its own mini-batch of data, therefore the {\em global batch size} is $N$-times the mini-batch size per device. Thus, $\frac{S_1}{S_N}$ is also equal to $N$. 

$E_1$ is the number of epochs required to converge when one device is used, while $E_N$ is the number of epochs required when $N$ devices are used. At larger global batch sizes (higher $N$), the gradients from a larger number of training samples are averaged which results in model over-fitting as well as a tendency to get attracted to local minima or saddle points. This eventually leads to poor generalization of the network~\cite{hoffer2017train, goyal_2017, smith_2017, Jastrzebski_2017, Li_2014, Keskar_2018} and therefore more epochs are typically required to converge. As such, $\frac{E_1}{E_N}$ is usually less than one. Equation~\ref{eq:speedup_DP_1} can thus be simplified as:

\vspace{-0.6cm}
\begin{equation}
\label{eq:speedup_DP}
\begin{split}
    SU_N &= SE_N \times N \times \frac{E_1}{E_N}
\end{split}
\end{equation}
\vspace{-0.6cm}

When training at larger device counts ($N$) both $SE_N$ and $\frac{E_1}{E_N}$ decrease. At large global batch sizes, hyper-parameter tuning (which is a challenging and time consuming task) can be used to try and minimize the increase in number of epochs required for convergence. However for any particular network, beyond a certain global batch size it has been often observed that the number of epochs required to converge increases rapidly, even with hyper-parameter tuning~\cite{goyal_2017}. We describe how we calculate the values for $SE_N$, ${E_1}$, and ${E_N}$ in detail in Section~\ref{sec:methodology}.
%Determining $SE_N$ for a given model (on a particular system) can be measured by exhaustively running the model on hardware or it can be calculated using analytical models to estimate computation time and communication overheads~\cite{Yan_2015,Das_2016}. To calculate $\frac{E_1}{E_N}$, a given model needs to be trained to the desired accuracy using global batch sizes corresponding to one device (for ${E_1}$) and $N$ devices (for ${E_N}$). More details regarding our own measurement of these factors is provided in Section~\ref{sec:methodology}.
\vspace{-0.1cm}
\subsection{Quantifying Model Parallel Training Time}
As shown in Figure~\ref{fig:model_parallel_training}, MP enables more than one device to work on the same mini-batch at the same time. This directly reduces the time taken for one training step; term $T$ in Equation~\ref{eq:training_speedup}. 
We call this speedup from M-way MP, $SU^M$ and it can be measured using real hardware by splitting a model across multiple devices and measuring per step execution time or estimated using a numerical model.
Note that the $SU^M$ speedup already includes the communication cost of data movement between dependent operations placed across multiple devices.

As previously noted, the global batch size does not increase when employing MP. Therefore, the number of steps per epoch (term $S$ in Equation~\ref{eq:training_speedup}) and number of epochs required to converge (term $E$ in Equation~\ref{eq:training_speedup}) do not change. As such, improving $SU^M$ reduces convergence time by solely reducing term $T$ in equation~\ref{eq:training_speedup} while the other two terms remain constant. We find that typically the inherent parallelism of a given model or its implementation, limits the achievable $SU^M$. As a result, {\em MP alone} is not been considered a broadly applicable scalable parallelization strategy. However, we show that MP {\em can be combined} with DP to extend training scalibility beyond today's limits.
%We will show that for three important DL models, MP {\em can be combined} with DP to extend training scalibility beyond today's limits.
%EIMAN-1-23-19: Changed the above sentence and took out the three important models because it seems like that's what the sentence is emphasizing, whereas its the combination that its emphasizing. They already know about the three networks from intro and will shortly hear about it in methodology too.

\subsection{Hybrid Data and Model Parallel Training:}
\label{sec:speedup_dp_mp}

In Section~\ref{sec:speedup_dp}, we introduced the speedup obtained by $N-way$ DP in Equation~\ref{eq:speedup_DP}.  Now, let's assume we have scaled our training system up to $N$ devices using $N-way$ DP and are happy with the training speedup achieved. If additional devices (say $M\times N$ devices, where $M$ is an integer) were to become available for training, how should we best use these devices for distributed training?  Our goal is to identify when to continue to use DP alone, and when to combine DP with MP to obtain the highest possible training speedup. Using DP alone, the speedup from $M\times N$ devices compared to one device is (substituting $M\times N$ for $N$ in Equation~\ref{eq:speedup_DP}):

\vspace{-0.4cm}
\begin{equation}
\label{eq:speedup_NM_DP}
\begin{split}
    SU_{M\times N} &= SE_{M\times N} \times M\times N \times \frac{E_1}{E_{M\times N}}
\end{split}
\end{equation}
\vspace{-0.4cm}

A few observations are important when comparing the speedup from $M\times N$-way DP (Equation~\ref{eq:speedup_NM_DP}) and speedup from $N$-way DP (Equation~\ref{eq:speedup_DP}): First, scaling efficiency is generally lower for the system with $M\times N$-way DP compared to $N-way$ DP~\cite{Thakur_2005, Patarasuk_2009}. This is because all-reduce communication happens between a larger number of devices. Depending on the values of $N$, $M$, and system configuration, all-reduce communication potentially crosses slower inter-node links that leads to increased all-reduce times and reduces $SE_{M\times N}$~\cite{Shi_2017, nccl}. 
Second, since global batch size is larger at $M\times N$ devices (to maintain a constant mini-batch size), the number of steps per epoch is smaller by a factor $M$ compared to $N$-way DP. Third, the number of epochs required, $E_{M\times N}$, is greater than or equal to $E_N$.  These factors all trend towards lower efficiency as the number of devices employed in DP training grows.

When using $M\times N$ devices in a hybrid parallelization strategy of $N$-way DP where each worker uses $M$-way MP, we consider each worker's per step speedup to be $SU^M$. Thus the overall training speedup can be expressed as:

\vspace{-0.4cm}
\begin{equation}
\label{eq:speedup_NM_MP_DP}
\begin{split}
    SU_{N}^{M} &= SU^M \times SE_{N} \times N \times \frac{E_1}{E_{N}}
\end{split}
\end{equation}
\vspace{-0.4cm}

When comparing hybrid $N$-way DP with $M$-way model parallel workers, versus $N$-way DP with single GPU workers, the global batch size will remain the same. This is because in the $M\times N$-device configuration, every $M$ devices are grouped into a single data-parallel worker. Thus, the number of steps per epoch remains the same as that of $N$-way DP at $N$ and $\frac{E_1}{E_N}$ remains unchanged as well. As such, the per-step speedup achieved through MP increases the overall training speedup by a factor of $SU^M$, when comparing Equations~\ref{eq:speedup_DP} and~\ref{eq:speedup_NM_MP_DP}.

\begin{figure}
    \centering
    \includegraphics[width=0.9\linewidth]{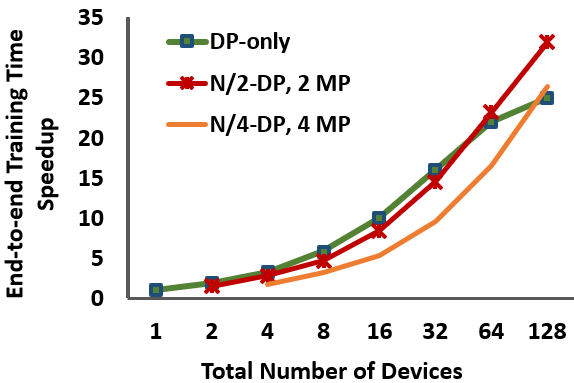}
    \caption{An example plot showing the speedup obtained from DP alone, and the hybrid strategy. $N$ refers to the total number of devices used for training.}
    \label{fig:high_level_concept}
\end{figure}
    
%\noindent
%\textbf{Choosing the best parallelization strategy:\\}
%\noindent
\vspace{-0.2cm}
\subsection{Choosing the Best Parallelization Strategy}
\label{sec:choosing_best_strategy}

By substituting Equations~\ref{eq:speedup_NM_DP} and~\ref{eq:speedup_NM_MP_DP} into Equation~\ref{eq:speedup_overall} we can determine the conditions under which using hybrid parallelization will be better than DP scaling alone. Equation~\ref{eq:speedup_overall} shows that if the speedup obtained from MP (for a given model parallelization step) is large enough to overcome the scaling and statistical efficiency loss that comes from increased communication, synchronization overhead, and global batch size increase respectively, employing a hybrid MP and DP strategy will improve network training time.

\vspace{-0.3cm}
\begin{align}
\label{eq:speedup_overall}
\begin{split}
    SU_{N}^{M} &> SU_{M\times N}
    \\
     SU^M \times SE_{N} \times N \times \frac{E_1}{E_{N}} &> SE_{M\times N} \times M\times N \times \frac{E_1}{E_{M\times N}}
     \\
     SU^M > M\times &\frac{SE_{M\times N}}{SE_N}\times \frac{E_N}{E_{M\times N}}
\end{split}
\end{align}
\vspace{-0.4cm}

Figure~\ref{fig:high_level_concept} illustrates this concept using a hypothetical scenario. Let's assume implementing MP provides a 45\% and 65\% improvement with two and four GPUs respectively. The DP-only strategy scales well up to 32 devices after which the improvement in speedup slows down. This enables a hybrid 32-way DP \& 2-way MP hybrid parallelization strategy to perform better than 64-way DP given the scaling and statistical efficiency losses at 64 devices, for this example. 

Similarly, a hybrid 16-way DP \& 4-way MP hybrid strategy outperforms DP-only when scaling from 32 to 128 
%this used to say 64 to 128, changed it to 32 to 128, because its 4-way MP
devices. However in this example, this hybrid strategy's performance is not as good as the hybrid strategy of 32-way DP \& 2-way MP. The reason is that 4-way MP's per step speedup ($SU^4$) does not overcome the trade-off (of using four machines for each data-parallel worker) as efficiently as 2-way MP's per step speedup, $SU^2$ (when using two machines for each data-parallel worker). Depending on these relative improvements at any device count, the choice of parallelization strategy is critical to the training speedup obtained when scaling to yet larger number of devices. This choice depends on the DL network's properties and system configuration parameters as described above, so there is no one size fits all solution to efficient scale-out multi-device training.

\section{Methodology}
\label{sec:methodology}
%\subsection{Benchmarks and Training Setup}

We use the following DL models in our evaluations with their default hyper-parameters, unless otherwise specified:

\vspace{-0.3cm}
\begin{itemize}
\setlength\itemsep{0em}
    \item {\em Inception-v3}~\cite{szegedy_2015} is used for image recognition and visual feature extraction. The network is composed of multiple blocks, each with several branches of convolution and pooling operations. These branches can be executed in parallel.  We use the implementation provided with the public NVIDIA Tensorflow container 18.07~\cite{tf_imagenet} and train the network using the Imagenet dataset~\cite{imagenet}. We scale the initial
    %EIMAN-1-23-19: SAPTADEEP TODO: Fix above reference to container.
    learning rate linearly with the increase in global batch size as originally proposed by Goyal et al.~\cite{goyal_2017}. For measuring epoch counts, we train the model until a training loss of 6.1 is achieved.
    
    \item {\em GNMT}~\cite{wu_2016} is a language translation network with attention mechanism~\cite{wu_2016, Bahdanau_2014}. We use 4 LSTM layers of size 1024 in the encoder and decoder. We use the public repository at~\cite{GNMT_v2} as the basis of our implementation. We use exponential learning rate warm-up for 200 training steps. The learning rate decay is started after 6000 steps and decays for a total of four times after every 500 iterations with a decay factor of 0.5. Such a technique has been shown to scale well when global batch size is scaled. We train the network using the WMT'16 German-English dataset ~\cite{guillou2016findings} until a BLEU score of 21.8 is achieved.
    
    \item {\em BigLSTM}~\cite{bigLSTM} is a large scale language modelling network. It consists of an input embedding layer of size 1024, 2 LSTM layers with hidden state size of 8192, and a Softmax projection layer of size 1024. We implemented the network in the public NVIDIA PyTorch container v19.06, used a learning rate of 0.1, and trained using the 1 billion word language modelling dataset %~\cite{chelba_2013} 
    to a perplexity of 67.
    
   % \item {\em DeepSpeech2} (DS2)~\cite{DS2} is an end-to-end speech recognition network. The network architecture comprises of multiple layers starting with one or more convolutional input layers, followed by five RNN layers, a look-ahead convolution layer and a fully connected output layer. The inputs to the network are a sequence of log spectogram of power normalized audio clips and outputs are the alphabets of a language.
    %EIMAN-1-13-19: The above sentence does not make sense to me. Please revise and simplify.
    %We use an optimized PyTorch implementation of the network. We train the network on the Librispeech dataset~\cite{librispeech}. We train the model to attain word error rate (WER) of XXXX.
    
\end{itemize}
\vspace{-0.3cm}

\subsection{System Configuration and Evaluation Points}

For our experiments, we use an NVIDIA DGX-1~\cite{DGX} with 4 Tesla V100 GPUs~\cite{V100} connected via NVLink ~\cite{NVLINK} with 16GB of memory capacity.  In the BigLSTM experiments we used a similar system but with GV100 cards having 32GB of memory, because this network requires more capacity to execute on a single GPU. We use NCCL2.0 based all-reduce communication for gradient sharing.

In order to project when hybrid training will perform better than DP alone, we need to measure the {\em epoch counts to convergence} and {\em scaling efficiency} for DP (defined in Section~\ref{sec:speedup_dp}) for different GPU counts. We also require the speedup achieved via MP when $M$ GPUs are used for a model-parallel worker in a hybrid strategy. 
%In our evaluations, we fix $M=2$ and answer the question: When we have scaled to $N$-way DP and want to scale up further, is the best speedup achieved with $2\times N$-way DP, or a hybrid strategy of 2-way MP workers in an $N$-way DP configuration? 
%EIMAN-1-23-19: The sentence above is redundant with what we have said before other than the M=2 part which is discussed in the following sentence. So removing the above sentence and massaging the one below a bit.
Without loss of generalization, we use $M=2$ for the DL models we use to make a case for future hybrid parallelization strategies. The value chosen for $M$ for an arbitrary DL model will always depend on the speedup obtained from M-way MP and slowdown in scaling efficiency the DP implementation incurs.

\subsection{Measuring Epoch Counts to Convergence}
\label{sec:measuring_epoch}
Typically, epoch counts to convergence for DP on $N$ compute nodes is obtained by running the training on $N$ nodes. We select mini-batch sizes to saturate single GPU throughput or lower if the desired mini-batch size is limited by GPU memory capacity. We perform experiments on a 4-GPU NVIDIA DGX system, so the maximum global batch size possible to measure is $4\times B$, where the mini-batch size is $B$.  To emulate larger global batch sizes (corresponding to more than four GPUs), we use the delayed gradient update approach~\cite{ott_2018} where multiple mini-batches are processed per GPU before the gradients are shared for weight update.  For example, to emulate a batch size of $16\times B$ that would be used in a 16 GPU system, each GPU runs the forward and backward propagation of four mini-batches before the GPUs share the gradients (using NCCL 2.0 based all-reduce~\cite{nccl}) and update weights.  This methodology allows us to measure the effect of global batch size on the epoch counts required for reaching a desired accuracy, at higher device counts than we have in our physical system. It is worth noting that even though we complete training of a DL model to find $E_N$, in practice, many DL models are often re-trained many times during development or as new data becomes available. Our proposed systematic modelling approach helps find the best parallelization strategy for optimizing the turnaround time of such subsequent training runs.

Learning rate schedules are sometimes optimized to keep epoch counts to convergence low at large global batch sizes. For example, the learning rate schedules we use for GNMT and Inception V3 were tuned accordingly for this purpose. However, in general, hyper-parameter tuning is time consuming and requires many training runs. Similar to prior work~\cite{goyal_2017}, we find that even with such tuning, beyond a certain global batch size, the number of epochs required to converge increases rapidly. As such, the proposals of this work are orthogonal to such efforts.

\begin{figure}
    \centering
    \includegraphics[width=0.85\linewidth]{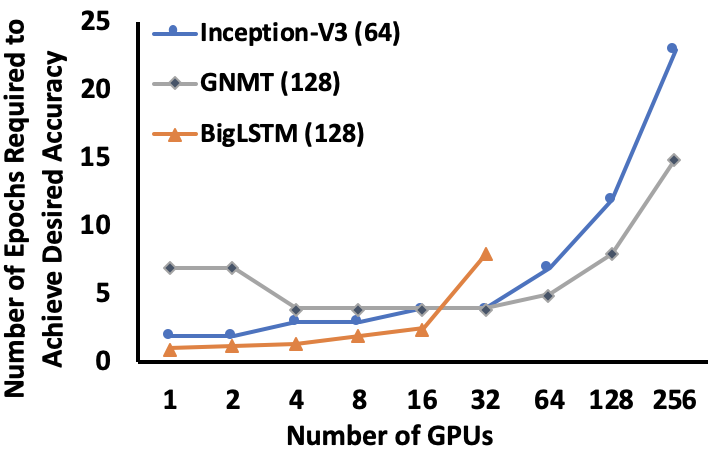}
    \caption{Number of epochs required for the networks to converge versus increasing global batch size with increase in the number of GPUs. We emulated larger global batch sizes corresponding to large number of GPUs using the technique described in Section~\ref{sec:measuring_epoch}}
    \label{fig:epoch}
\end{figure}

\begin{figure*}[t]
    \centering
    \begin{subfigure}{0.33\linewidth}
        \centering
        \includegraphics[width=\linewidth]{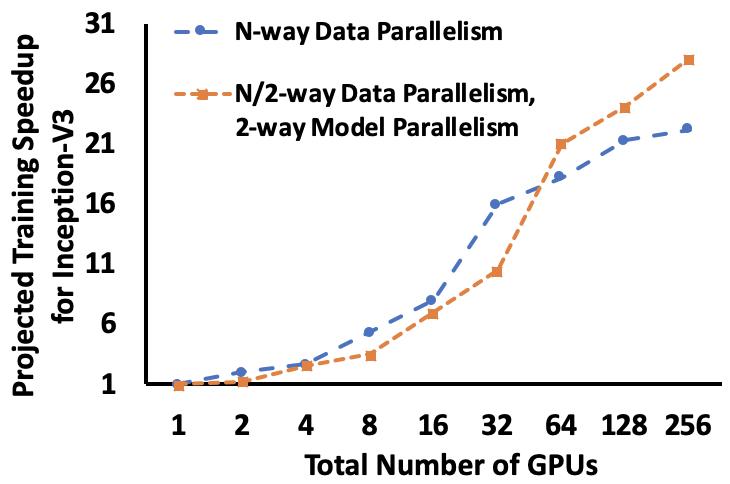}
        \caption{Inception-V3}
        \label{fig:speedup_inception}
    \end{subfigure}
    \begin{subfigure}{0.33\linewidth}
        \centering
        \includegraphics[width=\linewidth]{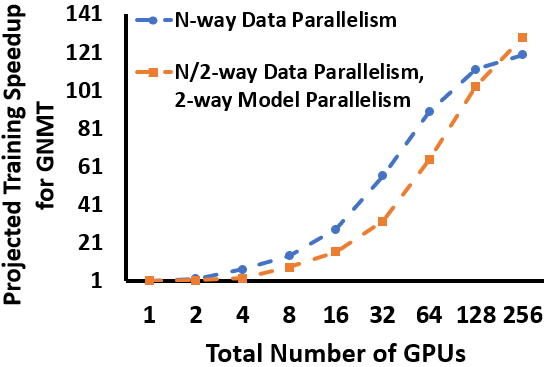}
        \caption{GNMT}
        \label{fig:speedup_gnmt}
    \end{subfigure}
    \begin{subfigure}{0.33\linewidth}
        \centering
        \includegraphics[width=\linewidth]{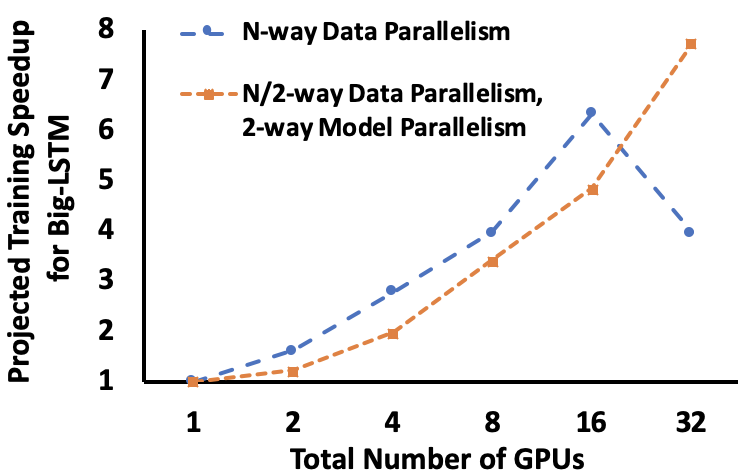}
        \caption{BigLSTM}
        \label{fig:speedup_biglstm}
    \end{subfigure}
    \caption{Projected speedup of hybrid MP-DP parallelization vs DP-only parallelization}
\end{figure*}

\subsection{Estimating Scaling Efficiency}
\label{sec:scaling_efficiency}

%Unlike the effect of global batch size on required epoch counts for reaching a desired accuracy,
Unlike the methodology we use for emulating larger global batch sizes than what our physical system allows, we can not obtain the scaling efficiency ($SE_N$) of data parallel training on larger number of GPUs, when using just four.  Thus, we conservatively assume a scaling efficiency ($SE_N$) of 1, i.e., the time overhead of communication and synchronization after each step is negligibly small compared to the time taken for the forward and backward passes. This optimistic assumption {\em minimizes the impact of hybrid parallelization}, but reflects the reality that framework developers are constantly working to improve overheads that hinder DP scaling efficiency. In fact for CNNs such as ResNet-50, relative scaling efficiency of $>$ 95\% has been achieved for 2048-way DP~\cite{Yamazaki_2019}.

\subsection{Model Parallel Splitting}
\label{sec:mp_splitting_implementation}

%As described in Section~\ref{sec:choosing_best_strategy}, the speedup obtained from MP is critical to the choice of when to employ hybrid parallelism. To implement MP for BigLSTM and GNMT, we split their DFGs across two GPUs using the pipeline parallelism technique described in~\cite{gpipe}. Exploiting pipeline parallelism is appropriate for 2-way MP on these networks due to the use of optimized libraries and fused RNN kernels in their implementations. Inception-V3's implementation allows a more traditional MP mapping of independent operations to different GPUs. We split Inception-V3's DFG across two GPUs using manual placement and also using an optimal operation-to-GPU mapping tool we developed for this purpose called DLPlacer. Additional details about DLPlacer are discussed in Section~\ref{sec:dlplacer}.

Inception-V3's implementation allows a traditional model parallel mapping of independent operations to different GPUs. As such, we split the model's DFG across two GPUs using DLPlacer, later described in Section~\ref{sec:dlplacer}. \textcolor{black}{We observed that beyond 2-way splitting of the Inception-V3 DFG, the MP speedup is marginal (see Figure~\ref{fig:mp_inception})}.
\textcolor{black}{For GNMT and BigLSTM, we split their DFGs using pipeline parallelism~\cite{gpipe}. Pipeline parallelism is appropriate for implementing MP on these networks due to the use of optimized libraries and fused RNN kernels in their implementations. Pipelining could similarly be useful for models which do not have parallel branches and are sequential in nature (e.g., ResNet, AmoebaNet).}

It is worth noting that the original GNMT implementation~\cite{wu_2016} uses 8-way MP. However, since we use a system with V100 GPUs that have 14x more FLOPs compared to the K80 GPUs used in that prior work, the ratio of communication overhead to computation is larger in our configuration. \textcolor{black}{We use up-to-date CuDNN libraries with fused RNN kernels and observe that splitting the model beyond 2-way provides marginal per-step speedup because of kernel overheads and pipeline imbalance.}

\section{Evaluation}\label{sec:eval}

Figure~\ref{fig:epoch} shows the number of epochs required to hit the desired accuracy versus the number of GPUs (workers) used in data parallel training. The number of epochs generally increases with an increasing number of GPUs (i.e., with increasing global batch size). For Inception-V3, the number of epochs increases sharply from four to seven as the global batch size increases beyond 2048 (i.e., 32 GPUs) and grows to 23 epochs at a global batch size of 16384 (i.e., 256 GPUs). For GNMT, the epoch count decreases slightly when going from two to four GPUs because the hyper-parameters used are tuned for large global batch sizes. Even with these tuned hyper-parameters, as the GPU count increases beyond 64, the number of epochs required grows rapidly. In BigLSTM, beyond 16 GPUs (i.e., global batch size of 2048), the number of epochs increases rapidly and in fact, 3.2 times the number of epochs is required for 32-way DP compared to 16-way DP. Beyond 32-way DP, training did not converge within a meaningful time limit. Overall, as we scale up the number of GPUs used in DP training, $\frac{E_1}{E_N}$ becomes smaller which ultimately hinders the overall speedup achievable through data parallel training alone. 

As described in Section~\ref{sec:end_to_end}, splitting each network across two GPUs using model parallelism results in per-step speedup when done successfully. Table~\ref{tab:MP_speedup} shows the measured MP speedups on our test system for our evaluated networks. Using the number of epochs required and per step speedup from MP, together with the conservative estimates of scaling efficiency, we can then calculate the minimum projected speedup (over DP-only) that can be obtained by implementing a hybrid parallelization strategy across different GPU counts. It is worth noting that the MP speedup achieved on Inception-V3 using expert manual placement of operations was 21\%. In Section~\ref{sec:dlplacer} we discuss DLPlacer, a tool we developed for optimizing operation-to-device placement, which improves the MP speedup for Inception-V3 to 32\%.
%It is worth noting that the MP speedup achieved on Inception-V3 using hand placement of operations was 21\%, versus the 32\% achieved using our DLPlacer tool; perhaps an indication of why DP vs MP tradeoffs have been historically undervalued when optimizing DL training.

\vspace{-0.2cm}
\paragraph{Inception-V3}
As shown in Figure~\ref{fig:speedup_inception}, beyond 32 GPUs, a hybrid parallelization strategy performs better than DP-only. This is because of the sharp increase in the number of epochs required when the global batch size grows beyond 2048 which saturates the speedup obtainable from DP-only parallelization. When moving from 32 GPUs to 64 GPUs, it is better to use the additional 32 GPUs to do 2-way MP, and our estimates show that the hybrid-strategy will outperform DP alone by at least 15.5\%. As the numbers of GPUs grow further, only marginal speedup can be obtained from DP-only parallelization and at 256 GPUs, the hybrid-strategy will be atleast 26.5\% better than the DP-only strategy.
\vspace{-0.2cm}
\paragraph{GNMT}
As shown in Figure~\ref{fig:speedup_gnmt}, GNMT scales very well to a large number of GPUs using DP alone. However, even with tuned hyper-parameters for larger batch sizes, DP-only speedup starts to slow down beyond 64 GPUs and dramatically slows down when moving from 128 to 256 GPUs. The hybrid parallelization strategy with 2-way MP and 128-way DP outperforms 256-way a DP strategy by 8\%. If the hyper-parameters would not have been tuned for large batch sizes, the gains from hybrid parallelism would be larger and the tipping point would occur at a lower number of GPUs.
\vspace{-0.2cm}
\paragraph{BigLSTM}
As shown in Figure~\ref{fig:speedup_biglstm}, beyond 16 GPUs, BigLSTM does not scale well with an increasing number of GPUs. This is because the statistical efficiency of training decreases rapidly with increasing global batch size, and therefore the significantly larger number of required epochs offsets the throughput increase of multiple GPUs. 
%The hybrid strategy performs within 4\% of the DP-only strategy at 8 and 16 GPUs. 
At 32-GPUs, the large loss in statistical efficiency impacts the overall training speedup of DP-only strategy and the speedup drops significantly. As a result, the hybrid policy provides a 1.22x speedup over the best performing scale of DP-only which happens at 16-GPUs, as Figure~\ref{fig:speedup_biglstm} shows.

\begin{table}[t]
\caption{MP splitting strategy and the speedup obtained when split across 2 GPUs}
\begin{tabular}{|c|c|c|}
\hline
\textbf{Network}   & \textbf{MP splitting strategy}   & \textbf{Speedup}  \\
\hline
Inception-V3 & Partitioned w/ DLPlacer & 1.32x    \\
\hline
GNMT & Pipeline Parallelism    &   1.15x \\
\hline
BigLSTM     & Pipeline Parallelism    & 1.22x    \\
\hline
\end{tabular}
\label{tab:MP_speedup}
\end{table}

In summary, these results show that when statistical efficiency loss reduces the effectiveness of DP-only parallelization, hybrid parallelization (combining DP with MP) will enable higher performance than employing DP alone. Notably, using real scaling efficiency loss values (we conservatively assumed $SE_N = 1$), the improvements from hybrid parallelization would be more pronounced since $\frac{SE_{2N}}{SE_N}$ is often smaller than 0.9 for large LSTM based networks. Based on Equation~\ref{eq:speedup_overall}, the smaller the ratio, the higher the speedup from hybrid parallelism ($SU_N^M$) compared to data-parallelism alone ($SU_{M \times N}$).

%In summary, these results show that for distributed training, when statistical efficiency loss reduces the effectiveness of DP-only parallelization, hybrid parallelization (combining DP with MP) will enable higher performance than employing DP alone. It is worth noting that if scaling efficiency loss (which we have not included in this evaluation due to our conservative assumption of $SE_N = 1$) were to also be considered, the improvements from hybrid parallelization would be more pronounced.

%In summary, for distributed training on a large number of GPUs, when statistical efficiency (epoch count) or scaling efficiency reduces the effectiveness of DP-only parallelization, this work demonstrates that hybrid parallelization (combining DP with MP) will enable higher performance than employing DP alone.

\section{Maximizing MP Performance}
\label{sec:dlplacer}
Maximizing the speedup obtained from MP for a given model improves the scalability of hybrid parallelism. For some networks, optimal placements are easy to achieve by examining a network's DFG. For others, finding the optimal operation-to-device placement that results in the maximum per-step speedup is non-trivial. To this end, we developed an integer-linear programming (ILP) based device placement tool called DLPlacer. DLPlacer maximizes resource utilization by extracting parallelism between operations in a model while also minimizing the communication overhead of moving data between the compute nodes. 

\begin{figure}
    \centering
    \includegraphics[width=0.85\linewidth]{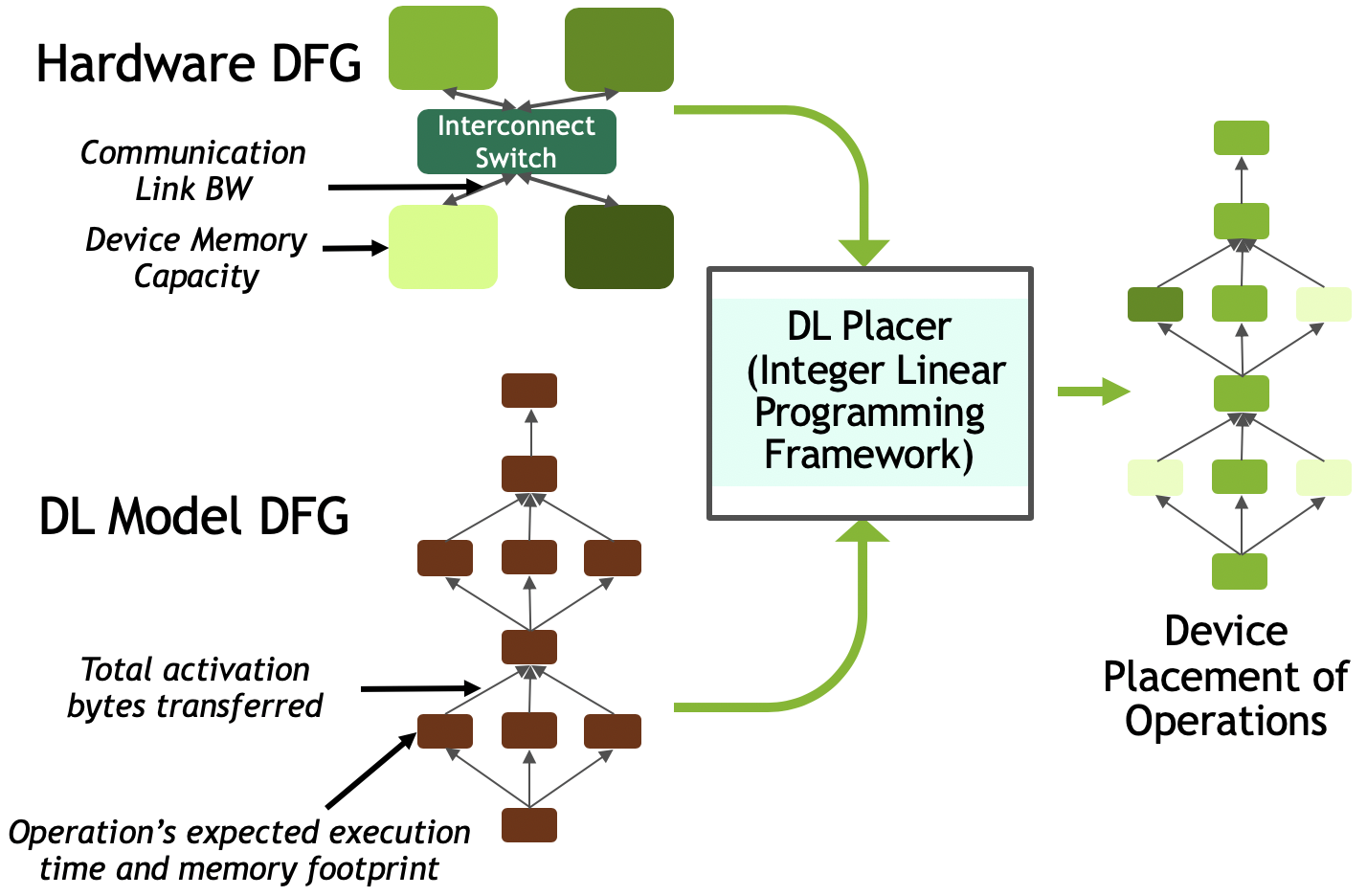}
    \caption{DLPLacer Flow Diagram.}
    \label{fig:dlplacer_flow}
\end{figure}

Figure~\ref{fig:dlplacer_flow} shows DLPlacer's tool flow. We express a DL model as a compute DFG, with a set of vertices $K$ corresponding to compute operations and a set of uni-directional edges $E$ showing operation dependencies. For example, for some $k1$, $k2$ $\in$ $K$ and $e_{k1, k2}$ $\in$ $E$, $e_{k1, k2} = 1$ means $k2$ is dependent on $k1$. The expected execution time of a vertex is represented as $\Delta(k)$ and the memory footprint of the vertex for a given batch size is represented as $M(k)$. Edge weight ($D(e)$) corresponds to the number of bytes exchanged between the operations it connects. The node and edge weights can be obtained by profiling a model on a compute device (e.g., GPU) or can be analytically calculated, with the former approach being more robust and the latter more flexible. 

Using similar notation, we express a system as a hardware graph~\cite{Nowatzki_2013_1, Nowatzki_2013_2} where a set of compute (e.g., GPUs) nodes $N$ and router nodes (network switches) $R$ are connected through a set of physical links $L$. As an example, for $n1$, $n2$ $\in$ $N$ and $l_{n1, n2}$ $\in$ $L$, $l_{n1, n2}$ = 1 means nodes $n1$ and $n2$ are connected. We assume physical links are bidirectional, so $l_{n1, n2}$ = $l_{n2, n1}$. The bandwidth of the physical link is denoted by $B(l)$.

DLPlacer's ILP solver minimizes per step training time by providing an assignment of compute DFG operations on to the hardware graph, a schedule, and a communication routing of activations, weights, and gradients. This is done by mapping model DFG's vertices to compute nodes, dependency edges to physical link mapping (if dependent vertices are placed in separate devices) and determining the execution start time of each vertex on a device. This mapping must satisfy a series of constraints and variables which are described next. A summary of all the variables is provided in Table~\ref{tab:notations} 

\begin{table}[h]
    \centering
    \scriptsize
    \caption{Summary of notations used in the ILP}
    \label{tab:notations}
    \begin{tabular}{|c|c|}
    \hline
    \bf{Notation} & \bf{Meaning} \\
    \hline
    \multicolumn{2}{|c|}{\textbf{Inputs : Computation DFG}}\\
    \hline
    $K$ & Set of compute operation/kernel vertices \\
    \hline
    $E$ & Set of edges between vertices \\
    \hline
    %G(V\cup E, V\cup E) & Compute DFG \\
    %\hline
    $\Delta(K)$ & Expected execution time of compute vertex \\
    \hline
    $M(K)$ & Memory footprint of the compute vertex \\
    \hline
    $D(E)$ & Number of data bytes transferred in an edge \\
    \hline
    \multicolumn{2}{|c|}{\textbf{Inputs : Hardware Graph}}\\
    \hline
    $N$ & Set of compute hardware nodes \\
    \hline
    $R$ & Set of router nodes \\
    \hline
    $L$ & Set of physical links connecting routers and hardware nodes \\
    \hline
    %$H(N\cup R\cup L$, & Directed graph describing the Hardware\\
    %$N\cup R \cup L)$ & \\
    %\hline
    $B(l)$ & Bandwidth of the physical links \\
    \hline
    $Mem(N)$ & Device memory capacity \\
    \hline
    \multicolumn{2}{|c|}{\textbf{Variables: Outputs}}\\
    \hline
    $P_{kn}(K,N)$ & Mapping of compute vertex to hardware node \\
    \hline
    $T_k(K)$ & Time a vertex is launched on the hardware \\
    \hline
    $C_{el}(E,L)$ & Mapping of dependency edges to physical links \\
    \hline
    \multicolumn{2}{|c|}{\textbf{Variables: Intermediate}}\\
    \hline
    $\Delta_{e}(L)$ & Delay of communication of edge $e$ \\
    \hline
    
    \hline
    \end{tabular}
    
\end{table}

Now, we describe the constraints in details.
%Any solution returned by DLPlacer must satisfy the valid mapping requirements of:

\paragraph{\bf{Placement of compute operation vertex:}}
If the binary variable $P_{kn}(k,n)$ = 1, then vertex $k$ is mapped to node $n$. Each operation of the compute DFG should be mapped to only one node on the hardware graph, therefore this gives us the following constraint:

\begin{equation}
    \begin{split}
        &~~~\forall{k}~~\sum_n P_{kn}(k,n) = 1
    \end{split}
\end{equation}

\paragraph{\bf{Routing of Activation Data:}}
The output from a vertex need to be routed to the dependent vertices through the physical communication links. Each edge $e$ needs to be mapped to a sequence of one or more links $l$. The path for communication must start from the origin vertex and end at the destination vertex if the origin and destination vertices are different. Therefore, for the source and destination nodes, exactly one link should be allotted for the edge, and for all other nodes (including router nodes). This constraint can be formulated as follows: 

\begin{equation}
    \begin{split}
        ~~~\forall{e,n, k_i, k_j} \mid &e_{k_i,k_j}=1~~if P_{kn}(k_i,n) != P_{kn}(k_j,n) \\
                                                    & \sum_{l|l_{n,nx}=1} C_{el}(e,l) = 1~~\forall nx \in N 
    \end{split}
\end{equation}

To find a contiguous path, we enforce for all non-source and non-destination nodes (includes router/switch nodes) that either two links should be allotted, one for the incoming traffic and one for the outgoing traffic or no links should be allotted.

\begin{equation}
    \begin{split}
     ~~~\forall & {e, k_i, k_j} \mid e_{k_i, k_j}=1, n \mid P_{kn}(k_i,n) = P_{kn}(k_j,n) = 0 \\
     & \sum_{l|l_{n,nx}=1} C_{el}(e,l) = \sum_{l|l_{n,ny}=1} C_{el}(e,l) \forall nx, ny \in N\cup R
    \end{split}
\end{equation}

\paragraph{\bf{Scheduling of Vertices:}}
Vertices need to be scheduled such that their dependencies are met. We calculate the time at which a vertex can begin executing by considering the start time of the other vertices it is dependent upon and the execution time and communication delay of the input activations. 

\begin{equation}
    \begin{split}
        ~~~\forall{k_{src}, k_{dest}, e} &\mid e_{k_{src}, k_{dest}}=1, \\
            T(k_{dest}) \geq T(k_{src}) &+ \Delta(k_{src}) + \Delta_e(L) 
    \end{split}
\end{equation}

This equation ensures that a vertex $k_{dest}$ can begin only after all the vertices it is dependent upon ($k_{src}$) has finished executing and the input activations have been communicated to the device where $k_{dest}$ is placed.

$\Delta_e(L)$ is the time to communication the edge data. The amount of data that need to be routed between two vertices is the amount of total output activation (dependent on mini-batch size). We assumed that the time for communication would depend on the number of links it need to traverse and the bandwidth and latency of these links. Therefore, $\Delta_e(L)$ can be computed as follows:

\begin{equation}
    \begin{split}
        ~~~\forall e \in E,
             \Delta_e(L) = \sum_{l\in L} C_{el}(e,l)*(D(e)/B(l)+L(l))
    \end{split}
\end{equation}

Another timing related constraint comes from the fact that multiple operations can be mapped to a device but co-located vertices cannot be scheduled on the same device at the same time. The start of the execution of consecutive operations on a device should atleast be separated by the execution time of the operation which starts earlier among the two. Note that this constraint is unnecessary for operations which lie on the dependency path of each other because of the previous constraint. Therefore, this constraint can be formulated as follows:

\begin{equation}
    \begin{split}
        ~~~\forall{k_x, k_y, n} &\mid P_{kn}(k_x,n)=P_{kn}(k_y,n)=1~and~e_{k_x, k_y} != 1, \\
            &if~T(k_x) > T (k_y): \\
            &~~~~T(k_x) \geq T(k_y) + \Delta(k_y) \\
            &else:\\
            &~~~~T(k_y) \geq T(k_x) + \Delta(k_x)
    \end{split}
\end{equation}
\vspace{-0.2cm}
\paragraph{\bf{Device memory capacity constraint:}}

This constraint ensures that the summation of the memory footprint of all the vertices placed on a device does not exceed the device memory capacity.

\begin{equation}
    \begin{split}
        ~~~\forall n \in N,~~
             Mem(n) \geq \sum_{k\in K} P_{kn}(k,n)*M(k)
    \end{split}
\end{equation}

\begin{figure*}[ht]
    \centering
    \includegraphics[width=0.95\linewidth]{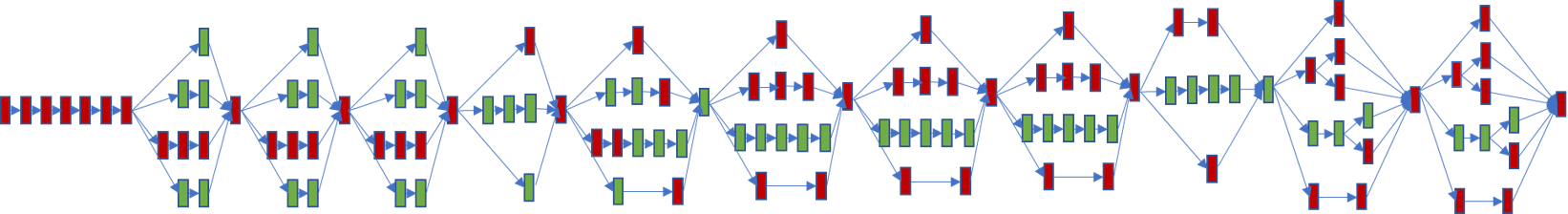}
    \caption{DLPlacer's placement solution for Inception-V3. Different colors denote different devices.}
    \label{fig:mp_inception_placement}
\end{figure*}

In this DLPlacer framework, we assumed the following:
\begin{enumerate}
\setlength\itemsep{0em}
\vspace{-0.05in}
    \item Two operations which are co-located on a device are executed back-to-back, without any delay in between the end of one operation and the beginning of the other.
    \item Communication of tensors between devices are overlapped with computation.
\end{enumerate}

Based on these constraints and assumptions DLPlacer predicts the training speedup for a given MP solution. In our work, we considered operations at the granularity of tensorflow operations (e.g., conv2D, conv3D), however DLPlacer can be used to even find placements when the operations are partitioned into finer granularity operations (e.g., partitioned by channels, filters etc.). But, such fine grained operation splitting requires framework support for correct back-propagation and therefore was not a focus of this work. Note that because of framework-induced overheads and unmodeled operating system effects, correct prediction of the exact speedup is difficult. Modelling these overheads is challenging and often depends on the mapping of kernels to high-level operations (e.g., mapping of CuDNN~\cite{cudnn} kernels to convolution/FC/etc.), device architecture, and the runtime implementation. Despite the challenges in accurate prediction, we believe ILP based MP optimization is worthwhile to pursue based on the observed improvements over manual optimization.

\noindent\textbf{Inception-V3 Case Study}
\label{sec:case_study_inception}

As inputs to DLPlacer, we analytically calculate the execution and communication times of the operations in the Inception-V3 DFG. For example, given the input/output tensor sizes of a convolution operation, we calculate the number of floating point operations (FLOPs) required, and based on advertised compute capability of NVIDIA's V100, we calculate the operations' expected execution time. Similarly, communication time between nodes is calculated based on the tensor sizes of the nodes in the model DFG along with NVLink bandwidth and latency. The placement solution of Inception-V3 using 2-GPUs is shown in Figure~\ref{fig:mp_inception_placement}. We implement the placement directives from DLPlacer using Tensorflow's $tf.device()$ command, and we have validated DLPlacer's speedup estimation against real hardware performance.

\begin{figure}
    \centering
    \includegraphics[width=0.95\linewidth]{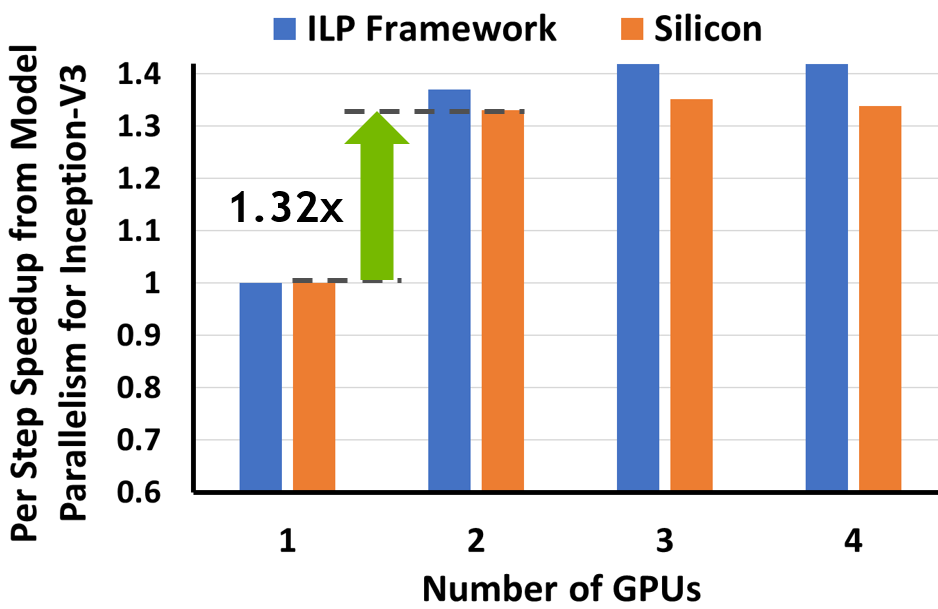}
    \caption{Normalized per-step speedup from model parallelism as estimated by DLPlacer and obtained from silicon experiments for the Inception-V3 network.}
    \label{fig:mp_inception}
\end{figure}

In Figure~\ref{fig:mp_inception}, the blue bars show the normalized per-step speedup estimated by DLPlacer for the optimal placement solution it finds. DLPlacer's runtime on an 18-core Xeon-E5 system to find Inception-V3's placement solution is $\sim$11-18 minutes depending on the number of device nodes in the hardware graph.
The orange bar for each configuration shows speedup as measured on real silicon with DLPlacer's placement applied to the Tensorflow implementation. The speedup-ups measured by DLPlacer are within 6\% of the actual speedup obtained from the silicon runs. It is interesting to note that the 1.32x speedup obtained with the real silicon 2-GPU placement is almost the same as what is optimally obtainable with three or four GPUs. This is due to the limited parallelism available in the network, which DLPlacer almost completely exploits with a 2-GPU placement. Identifying a 2-GPU placement that gives this performance by simple observation of the network and without using a tool like DLPlacer is non-trivial. DLPlacer essentially finds placement with the shortest possible critical path among many feasible placement solutions and places the operations on the critical path in one GPU so as to avoid communication overhead. This shows the importance of such a tool for maximizing performance obtainable from MP while using minimum number of GPUs.
%It is interesting to note that with 2-GPU MP, the speedup obtained is about 1.32x compared to the single GPU case, where we find that increasing the number of GPUs to three or four does not provide any further significant speedup. This is because the parallelism that can be extracted from the model is limited to what an optimal 2-GPUs placement can achieve.

%Even though close correlation between DLPlacer (with analytical model driving the DL input DFG) and silicon %runs may not exist for all networks and frameworks, we note that DLPlacer is not the essence of the broader %idea discussed here. DLPlacer is one option to help with obtaining the highest performance possible from the %MP implementation of a model to make the hybrid MP and DP parallelization strategy perform as best as it can. %We discuss other relevant options in Section~\ref{sec:related_work}.  

\section{Related Work}
\label{sec:related_work}
This work identifies scaling and statistical efficiency losses as the largest challenges to scalable data parallel training, but researchers are improving the scalability of both data and model parallel training rapidly. We summarize the most significant related advancements here.

\textcolor{black}{\subsection{Hybrid Parallelization}
Previous work~\cite{Das_2016, Yadan_13, srinivas_2018} has also used hybrid parallelization for scaling DL training. To the best of our knowledge, none of these proposals provides a systematic method to identify which strategy is best for scaling-out network training at different device counts. 
Das et al.~\cite{Das_2016} perform hybrid training on CPUs and maintain the global batch size by shrinking the mini-batch size per CPU, but do not incur a statistical efficiency loss because a small mini-batch size is large enough to saturate CPU throughput.  Maintaining a constant global batch size while shrinking the mini-batch size (per compute device) can also be done for GPUs, however GPUs typically require larger mini-batch sizes to maintain high utilization.  
Yadan et al.~\cite{Yadan_13} show that a hybrid (2-way DP, 2-MP) approach performs better than both MP-only and DP-only when training AlexNet on a 4-GPU system, but do not discuss the cause of the results or evaluate this effect across different GPU counts.
Dean et al.~\cite{Jeff_2012} used hybrid parallelism to train models which would not fit in a single GPU's memory. Therefore, in each data parallel worker, the model replica is model parallelized across multiple devices. However with increase in capacity of memory capacity in today's GPUs, large models such as Inception-V3, GNMT etc. can fit in to a single GPU memory while using sufficiently large mini-batch size to saturate the compute throughput. Moreover, using model parallelism for models that do not fit in a single GPU’s memory is largely orthogonal to the issue we address in this work.
Amir et al. ~\cite{Amir_2017} have shown that hybrid parallelization strategy can result in lower communication overhead over both MP and DP. None of these works, however, have provided a systematic analysis of finding what parallelization strategy would minimize the end-to-end training time when a set of $N$ compute devices are available for training. Moreover, implementing hybrid parallelism is often tricky because finding the optimal strategy to split a model is non-trivial and is dependent on the model DFG and system hardware.}

\subsection{Orthogonal Parallelization Strategies}
Exploiting model parallelism is just one way to achieve per step speedup without increasing global batch size. Other strategies exist that can be combined with, or used in place of, model parallelism to augment data parallel scaling under our proposed model.  Jia et al.~\yrcite{jia_2018} propose layer-wise parallelism for CNNs where each network layer can use an individual parallelization strategy. 
A combination of the 4D tensor dimensions can be used to parallelize a given layer and exploring multiple dimensions may provide larger runtime benefits than MP. However, such a technique is not yet supported by most frameworks and is evaluated using a custom framework (Legion~\cite{Bauer_2012}).
Similar to GPipe~\cite{gpipe} (discussed in Section~\ref{sec:background}), Harlap et al.~\yrcite{pipedream} propose partitioning a DL model's DFG into multi-layer stages and applying pipeline parallelism.
To enable maximum device utilization, PipeDream uses asynchronous weight updates
which can lead to poor statistical efficiency as the number of devices increases. It is likely that one or a combination of the layer-wise, pipeline, and model parallelism techniques can be combined with DP training to maximize end to end training performance and efficiency.

\subsection{Alternate Techniques to Improve DP Scaling}
Data parallel training employing sync-SGD suffers from poor scaling efficiency due to synchronization overheads. Prior work~\cite{async_sgd1, async_sgd2, async_sgd3, async_sgd4} has attempted to address this
by using asynchronous SGD. However, asynchronous SGD can still result in poor statistical efficiency while making performance debugging difficult. Hyper-parameter tuning is a broad approach to improving statistical accuracy and training convergence. Techniques such as tuning and scaling learning rates~\cite{goyal_2017, weird_trick, yang_2017, smith_2017, Jastrzebski_2017, hoffer2017train}, or auto-tuning the momentum~\cite{zhang2017yellowfin} are several important examples.
However, these techniques are very problem specific, require extensive knowledge of the DL models, and are very time consuming for developers. Furthermore, hyper-parameter tuning is not always effective~\cite{shaulle_2018}. 

Other works~\cite{Polyak_1992, koliousis2019crossbow} propose using a different learning algorithm, called model averaging, for training with small batches. An average model can asymptotically converge faster, but finding the asymptotic region is difficult~\cite{xu_2018}. Koliousis et al.~\yrcite{koliousis2019crossbow} use multiple learners (each using a small batch size) run on many GPUs, and an average main model is used to synchronously track the learning. Model averaging is not yet mainstream or supported by popular DL frameworks and thus requires custom re-implementation of the DL models.

\subsection{Reinforcement Learning-based Device Placement}
Prior work has shown that by using reinforcement learning-based (RL-based) placement of operations onto devices, MP can achieve training speedup and that the RL generated placement is non-trivial~\cite{mirhoseini2018a}. However, RL-based approaches can be long-running and compute-intensive with no notion of optimality. On the other hand, DLPlacer can provide optimal device placement solutions, though can still be compute intensive for complex DFGs and when system graph contains a large number of devices. \textcolor{black}{However, it should be noted that for simpler DFGs, simple heuristics could achieve near-optimal placement results.}

\subsection{Framework Support}
As we discuss in Section~\ref{sec:mp_splitting_implementation}, we implement MP differently for our BigSLTM and GNMT evaluation compared to Inception-V3. This is mostly driven by the baseline implementations of BigLSTM and GNMT, which makes it very non-trivial to exploit intra-layer MP in these networks. We use pipeline parallelism for exploiting inter-layer MP for these two models. While a given network's implementation can be one hurdle to exploiting MP, the framework it is implemented with can also add to the complexity. TensorFlow and Pytorch have different levels of support for assigning operations or tensors to different devices, but neither provide any automatic intra-layer parallelism extraction support. DSSTNE~\cite{dsstne}, is Amazon's deep scalable sparse tensor network framework which has more complete support for extracting intra-layer parallelism. However, it only supports fully connected layers and therefore is not a versatile framework for implementing different types of DL networks such as CNN and RNN based networks. Also, this framework is not broadly used and as such was not a focus of our evaluation in this work.

\section{Conclusion}
This paper demonstrates the benefits of combining model-parallelism (MP) with data-parallelism (DP) to overcome the inherent scaling and statistical efficiency losses that data-parallel training has at scale. We analyze the end-to-end training time of DP to understand how scaling and statistical efficiency loss impacts training scalability, and show that the MP speedup achieved for a given DL model is critical to the overall scalability of a hybrid parallelization strategy. We demonstrate that when the global batch size in DP grows to a point where DP-only training speedup drops off significantly, MP can be used in conjunction with DP to continue improving training times beyond what DP can achieve alone. We evaluate the performance benefits of such a hybrid strategy and project that for Inception-V3, GNMT, and BigLSTM, the hybrid strategy provides an end-to-end training speedup of at least 26.5\%, 8\%, and 22\% respectively compared to what DP alone can achieve at scale.

\bibliography{ref}
\bibliographystyle{icml2019}
%%%%%%%%%%%%%%%%%%%%%%%%%%%%%%%%%%%%

\end{document}